\documentclass[a4paper, 10pt, conference]{ieeeconf}      

\IEEEoverridecommandlockouts                              

\usepackage{amsmath} 
\usepackage{amssymb}  

\usepackage{graphicx}
\usepackage{url}
\usepackage{makecell}
\usepackage{booktabs}
\usepackage[hidelinks]{hyperref}

\newcommand{\figref}[1]{{Fig.~\ref{#1}}}
\newcommand{\tabref}[1]{{Table~\ref{#1}}}

\title{\LARGE \bf
Evaluating Tactile Foundation Models for End-to-End Imitation Learning in Contact-Rich Manipulation
}

\author{
Masaki Murooka$^{1,2}$, Ryoichi Nakajo$^{1}$, Tomohiro Motoda$^{1}$, Koshi Makihara$^{1}$,\\ Hanbit Oh$^{1}$, Keisuke Shirai$^{1}$, Yukiyasu Domae$^{1}$
\thanks{$^{1}$Artificial Intelligence Research Center,
National Institute of Advanced Industrial Science and Technology (AIST),
2-3-26 Aomi, Koto-ku, Tokyo 135-0064, Japan.
{\tt\small m-murooka@aist.go.jp}}%
\thanks{$^{2}$CNRS-AIST JRL (Joint Robotics Laboratory), IRL,
1-1-1 Umezono, Tsukuba, Ibaraki 305-8560, Japan.}%
\thanks{This work was partially supported by JST CREST, Japan, under Grant Number JPMJCR2553, and JST Moonshot R\&D Program, Japan, under Grant Number JPMJMS263D.}%
}

\begin{document}

\maketitle
\thispagestyle{empty}
\pagestyle{empty}


\begin{abstract}
Tactile sensing plays an important role in imitation learning for contact-rich manipulation by providing direct observations of physical contact with objects.
Recently, several tactile foundation models (TFMs) have been proposed to learn general-purpose tactile representations through pretraining on large-scale tactile datasets.
However, how these pretrained tactile representations should be leveraged for end-to-end imitation learning remains insufficiently explored.
In this work, we integrate multiple TFMs into Diffusion Policy and systematically evaluate their effectiveness for contact-rich manipulation under common experimental settings.
We further investigate adaptation strategies, including encoder fine-tuning and task-specific mid-training, to understand how pretrained tactile representations can be effectively used for downstream imitation learning.
Through extensive evaluation across diverse manipulation tasks in both real-world and simulation environments, including more than 5,000 real-world policy rollouts, we provide empirical insights into the use of TFMs for end-to-end imitation learning.
Project page: \url{https://mmurooka.github.io/tfm4il-project-page/}.
\end{abstract}

\section{Introduction} \label{sec:introduction}

Robots are expected to autonomously perform manipulation tasks in a wide range of environments, from daily life to industrial settings.
Recent advances in imitation learning have enabled end-to-end policies to be learned from expert demonstrations.
In particular, visuomotor policies that take visual observations as input have achieved high performance across various manipulation tasks.

However, in contact-rich manipulation, visual observations alone are often insufficient to capture physical interactions between the robot and objects~\cite{TactileIL:Tsuji:IJRR2026}.
Tactile sensing is therefore particularly important, as it provides direct information about contact states and changes in interaction forces.
Although a growing body of work has explored end-to-end policy learning with tactile observations, the methodology remains less mature than that of visuomotor policy learning.
In particular, how tactile observations should be represented and incorporated into a policy has not yet been fully explored.

Recently, several tactile foundation models (TFMs) have been proposed to learn general-purpose tactile representations through self-supervised pretraining on large-scale tactile datasets~\cite{Sparsh:Higuera:CoRL2024,AnyTouch:Feng:ICLR2025,T3:Zhao:CoRL2024}.
Pretrained tactile representations have also been applied to end-to-end policy learning~\cite{PolyTouch:Zhao:ICRA2025}.
However, different TFMs have not yet been systematically evaluated under a common policy architecture and experimental setting.
The effect of different pretrained tactile representations on downstream policy learning therefore remains unclear.
Furthermore, the use of these representations for downstream tasks, including encoder fine-tuning and task-specific mid-training on demonstration data, has not been sufficiently investigated.

In this work, we systematically investigate the effectiveness of TFMs for end-to-end imitation learning in contact-rich manipulation by integrating multiple TFMs into a common Diffusion Policy architecture~\cite{DP:Chi:IJRR2025}.
Using a robotic manipulator equipped with optical tactile sensors on its gripper, we evaluate multiple tactile representations and training strategies under consistent experimental conditions and analyze their characteristics.
Our real-world evaluation comprises more than 5,000 policy rollouts.
We further conduct simulation experiments in a comparable setup that includes simulated optical tactile sensing.
Through these extensive experiments, we provide practical insights into the design choices involved in leveraging TFMs for imitation learning.





\begin{figure}[tpb]
  \centering
  \includegraphics[width=1.0\columnwidth]{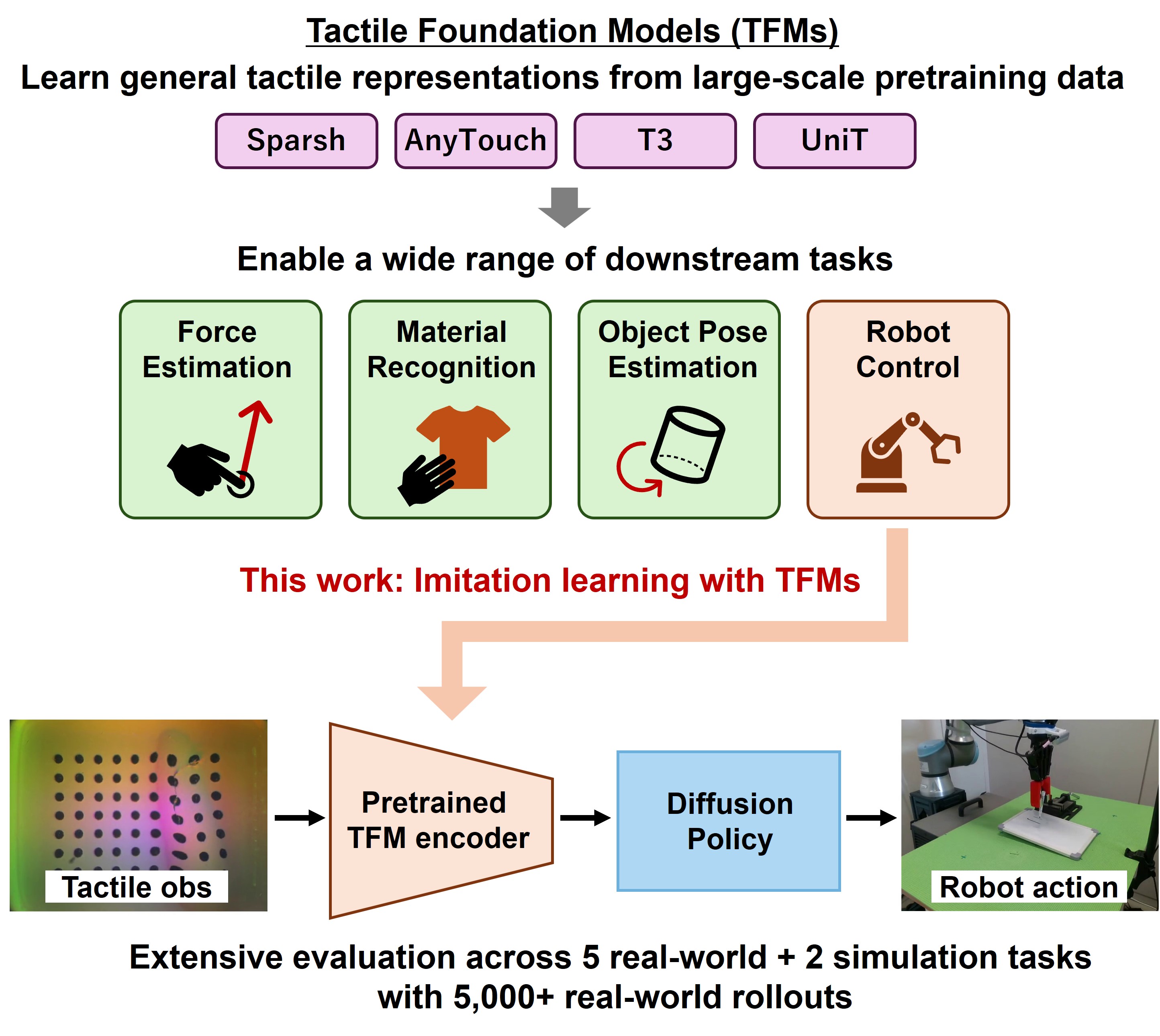}
  \caption{Imitation learning with tactile foundation models.}
  \label{fig:intro}
\end{figure}

The main contributions of this work are as follows:
\begin{itemize}
    \item We develop an end-to-end imitation learning framework that integrates multiple TFMs with Diffusion Policy and apply it to contact-rich manipulation in both real-world and simulation environments.
    \item Through extensive comparisons, we show that tactile observations improve performance over camera observations, while the benefits of pretrained TFMs vary across models and tasks.
    \item We show that fine-tuning a pretrained tactile encoder can improve performance over freezing it, whereas task-specific mid-training has task-dependent effects.
\end{itemize}

To facilitate reproducibility, we release the code for training and evaluation, 3D-printable models for mounting the tactile sensors, and a robot simulation environment with tactile sensing.



\section{Related Works} \label{sec:related-works}

\subsection{Tactile Sensing in End-to-End Policy Learning}

Imitation learning for robot manipulation has primarily focused on mapping visual observations to actions~\cite{ImitationSurvey:Fang:IJIRA2019,Robomimic:Mandlekar:CoRL2021}.
With the growing recognition of the importance of tactile sensing for contact-rich manipulation, recent studies have increasingly explored end-to-end policy learning with tactile observations.

A wide variety of tactile sensors have been used, ranging from optical tactile sensors~\cite{TactileAloha:Gu:RAL2025,exUMI:Xu:CoRL2025,TactileDP:Helmut:arXiv2025,UniT:Xu:RAL2025} to grid-based sensors~\cite{DexSkin:Wistreich:CoRL2025,TACT:Murooka:RAL2025,3D-ViTac:Huang:CoRL2024,SARNN:Ichiwara:ICRA2022,MultiFingerTactile:Funabashi:RAL2022}.
Their configurations also vary, including sensors mounted on parallel grippers~\cite{3D-ViTac:Huang:CoRL2024,TactileAloha:Gu:RAL2025,SARNN:Ichiwara:ICRA2022}, multi-fingered hands~\cite{DexSkin:Wistreich:CoRL2025,MultiFingerTactile:Funabashi:RAL2022}, and whole robot bodies~\cite{Punyo:Barreiros:ScienceRobotics2025,TACT:Murooka:RAL2025}.
While some studies use commercially available tactile sensors~\cite{TactileAloha:Gu:RAL2025,TACT:Murooka:RAL2025}, others develop custom sensors tailored to specific tasks or robotic platforms~\cite{Punyo:Barreiros:ScienceRobotics2025,DexSkin:Wistreich:CoRL2025,3D-ViTac:Huang:CoRL2024,MultiFingerTactile:Funabashi:RAL2022,PolyTouch:Zhao:ICRA2025}.

Various approaches have also been proposed to integrate tactile information into policies.
Tactile observations can be directly incorporated as additional inputs to existing policy architectures such as ACT~\cite{ACT:Zhao:RSS2023,TactileAloha:Gu:RAL2025,TACT:Murooka:RAL2025}.
Other approaches include aligning tactile information with 3D point clouds~\cite{3D-ViTac:Huang:CoRL2024}, jointly learning to predict tactile observations~\cite{exUMI:Xu:CoRL2025,SARNN:Ichiwara:ICRA2022}, and using graph neural networks that account for the spatial arrangement of tactile sensor cells~\cite{MultiFingerTactile:Funabashi:RAL2022}.

Simulation environments capable of reproducing tactile observations have also been developed~\cite{TACTO:Wang:RAL2022,TactileGym2:Lin:RAL2022} and used for tactile-based reinforcement learning~\cite{Punyo:Barreiros:ScienceRobotics2025,BiTouch:Lin:RAL2023}.

Despite the growing body of work on end-to-end policy learning with tactile sensing, the sensor setup, tactile representation, policy architecture, and target tasks vary substantially across studies.
This makes it difficult to isolate and compare the effect of tactile representations on policy performance.
In this work, we compare multiple pretrained tactile representations using a common sensor setup and policy architecture, focusing on their impact on end-to-end imitation learning.

\subsection{Tactile Foundation Models}

Tactile sensing has been used not only for end-to-end policy learning but also for various tasks such as contact force estimation~\cite{ForceEstimation:Chelly:IROS2025}, object pose estimation~\cite{Tac2Pose:Bauza:IJRR2023}, and material recognition~\cite{Octopi:Yu:RSS2024}.
To support such diverse downstream tasks, TFMs have recently been proposed to learn general-purpose representations from large-scale tactile data.

Representative TFMs for optical tactile sensors~\cite{OpticalTactileSurvey:He:TIM2025} include Sparsh~\cite{Sparsh:Higuera:CoRL2024}, AnyTouch~\cite{AnyTouch:Feng:ICLR2025}, and T3~\cite{T3:Zhao:CoRL2024}.
These models use self-supervised learning on tactile images collected across diverse environments, sensors, and contact conditions to learn representations transferable to multiple downstream tasks.
UniT~\cite{UniT:Xu:RAL2025} learns tactile representations using a VQGAN, aiming to achieve effective transfer with limited pretraining data.

Recent studies have also explored representations shared across different types of tactile sensors.
ContrastTouch~\cite{ContrastTouch:Rodriguez:ICRA2025} learns a shared representation across optical and soft-bubble tactile sensors, while UniTac-NV~\cite{UniTacNV:Hou:IROS2025} learns unified representations for non-optical grid-based tactile sensors.

Beyond perception and estimation tasks, TFMs have also been applied to single-task robot manipulation policies, demonstrating the use of pretrained tactile representations for end-to-end policy learning~\cite{Sparsh:Higuera:CoRL2024,AnyTouch:Feng:ICLR2025,T3:Zhao:CoRL2024,UniT:Xu:RAL2025}.
Pretrained TFMs have also been transferred to different tactile sensors and policy architectures~\cite{PolyTouch:Zhao:ICRA2025}.

However, these studies primarily focus on proposing individual TFMs and demonstrating their transfer performance, rather than comparing different pretrained tactile representations under common sensor, policy, and experimental settings.
In addition, how pretrained tactile representations should be adapted to downstream imitation learning, through approaches such as encoder fine-tuning and task-specific mid-training, has not been systematically investigated.
In this work, we integrate multiple TFMs into a common Diffusion Policy architecture and evaluate their effectiveness and adaptation strategies for contact-rich manipulation under consistent experimental conditions.

\begin{figure*}[tpb]
  \centering
  \includegraphics[width=1.7\columnwidth]{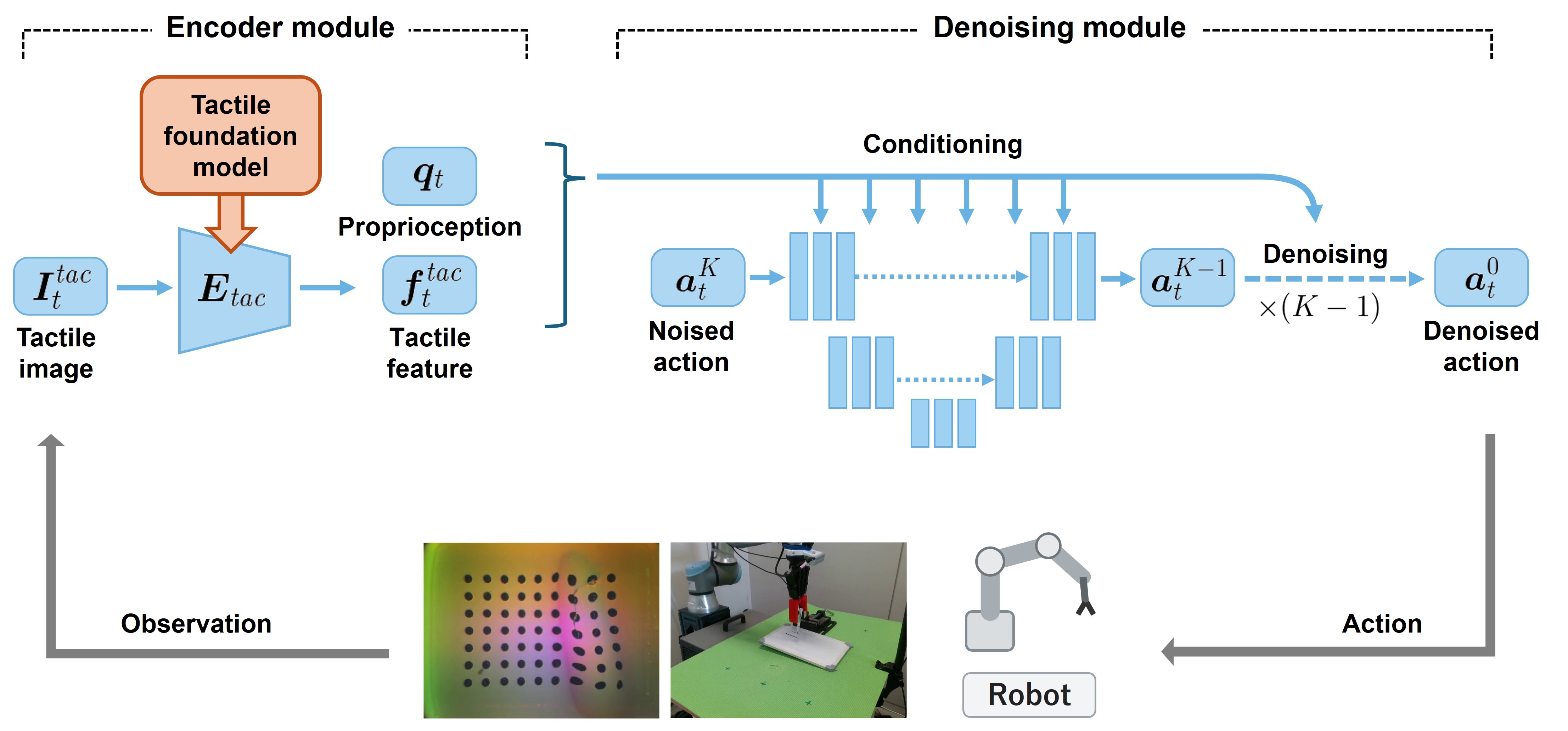}
  \caption{Overview of the policy architecture with a pretrained tactile encoder.}
  \label{fig:policy}
\end{figure*}

\section{Imitation Learning with Tactile \\Foundation Models} \label{sec:method}

This section describes the end-to-end imitation learning framework with TFMs evaluated in this work.


\subsection{Problem Formulation}

We consider an imitation learning policy that predicts a sequence of future actions from a sequence of observations:
\begin{equation}
  a_{t:t+T_a} = \pi(o_{t-T_o:t}),
\end{equation}
where $T_o$ and $T_a$ denote the observation and action horizons, respectively.

At each time step, the observation is defined as $o_t=(q_t,I_t)$, where $q_t$ denotes the robot proprioceptive state and $I_t$ denotes an image observation.
The image observation is encoded into a feature representation and concatenated with the proprioceptive state as input to the policy.
For $I_t$, we use either a camera image from a hand-mounted camera or a tactile image from an optical tactile sensor.

When tactile images are used, tactile features are extracted by a tactile encoder $E_{\mathrm{tac}}$:
\begin{equation}
  f_t^{\mathrm{tac}} = E_{\mathrm{tac}}(I_t^{\mathrm{tac}}).
\end{equation}
In this work, we use encoders from different TFMs as $E_{\mathrm{tac}}$ and evaluate their effectiveness for end-to-end imitation learning.




\subsection{Policy Architecture}

The overall policy architecture used in this work is shown in \figref{fig:policy}.
We use Diffusion Policy~\cite{DP:Chi:IJRR2025} with a convolutional U-Net as the common policy architecture.
The policy consists of an encoder module and a denoising module.

The denoising module generates an action sequence by iteratively denoising a noisy action sequence conditioned on observation features.
The encoder module extracts features from the robot proprioception $q_t$ and image observation $I_t$ to provide the conditioning for the denoising module.

The original Diffusion Policy uses ResNet-18~\cite{ResNet:He:CVPR2016} to extract features from RGB images and concatenates them with the proprioceptive state.
Although tactile and RGB images encode different physical information, this does not imply that image-based architectures such as ResNet are inherently unsuitable for tactile inputs.
Our focus is instead on the effect of tactile-specific pretraining, so we use pretrained TFM encoders as the tactile encoder $E_{\mathrm{tac}}$ to extract tactile features $f_t^{\mathrm{tac}}$ from tactile images $I_t^{\mathrm{tac}}$.




\subsection{Tactile Foundation Models} \label{sec:method-tfm}

We evaluate pretrained encoders from Sparsh~\cite{Sparsh:Higuera:CoRL2024}, AnyTouch~\cite{AnyTouch:Feng:ICLR2025}, T3~\cite{T3:Zhao:CoRL2024}, and UniT~\cite{UniT:Xu:RAL2025} as the tactile encoder $E_{\mathrm{tac}}$.
These models differ in their pretraining data, objectives, and architectures, but all provide latent representations of tactile images for downstream tasks.
We integrate each pretrained encoder into Diffusion Policy and apply model-specific pooling followed by a learnable linear projection to obtain a common 64-dimensional tactile feature.

\textbf{Sparsh}~\cite{Sparsh:Higuera:CoRL2024} learns general-purpose tactile representations from more than 460k tactile images collected with multiple optical tactile sensors.
It provides ViT-based encoders pretrained with different self-supervised learning methods.
We evaluate four variants: SparshMAE, SparshDINO, SparshDINOv2, and SparshIJEPA.

\textbf{AnyTouch}~\cite{AnyTouch:Feng:ICLR2025} learns sensor-agnostic tactile representations from data collected with multiple optical tactile sensors.
It combines masked modeling with multimodal alignment and cross-sensor matching.
We use its pretrained touch encoder as $E_{\mathrm{tac}}$.

\textbf{T3}~\cite{T3:Zhao:CoRL2024} is a Transformer-based framework trained on the Foundation Tactile (FoTa) dataset, which contains more than three million data points from 13 sensors and 11 tasks.
Its architecture consists of sensor-specific encoders, a shared trunk, and task-specific decoders.
We use the pretrained sensor-specific encoder corresponding to our tactile sensor together with the shared trunk.

\textbf{UniT}~\cite{UniT:Xu:RAL2025} learns a compact tactile representation using a VQGAN.
Unlike the other models, UniT does not rely on large-scale diverse pretraining data, but learns transferable representations from a small dataset collected with a single simple object.
Although it differs from typical large-scale TFMs, we include it as a pretrained tactile representation for comparison.
We use its pretrained CNN encoder as $E_{\mathrm{tac}}$.

\subsection{Fine-Tuning Strategies} \label{sec:method-adaptation}

We consider three training strategies for adapting pretrained tactile representations to downstream imitation learning: from-scratch training, freezing, and fine-tuning.
In \textbf{FromScratch}, the tactile encoder is trained together with the policy from random initialization without using pretrained weights.
In \textbf{Freeze}, the pretrained encoder is fixed, while the projection layer and Diffusion Policy are trained.
In \textbf{FineTune}, the pretrained weights are used for initialization, and the entire policy, including the tactile encoder, is updated end-to-end with the imitation learning objective.
Through these strategies, we examine the roles of tactile pretraining and downstream adaptation in policy learning.


\subsection{Task-Specific Mid-Training} \label{sec:method-mid-training}

To further adapt a pretrained tactile representation to a downstream task, we introduce task-specific mid-training between foundation model pretraining and imitation learning.
Mid-training uses demonstration data from the target manipulation task to learn a latent dynamics model that predicts future tactile features from consecutive tactile observations and a robot action:
\begin{equation}
  \hat{f}_{t+1}^{\mathrm{tac}}
  =
  F_{\mathrm{dyn}}
  \left(
    f_{t-1}^{\mathrm{tac}},
    f_t^{\mathrm{tac}},
    a_t
  \right),
  \quad
  f_t^{\mathrm{tac}}
  =
  E_{\mathrm{tac}}(I_t^{\mathrm{tac}}).
\end{equation}
Here, $F_{\mathrm{dyn}}$ is an MLP-based latent dynamics model.
The target tactile feature is extracted from the tactile observation at the next time step using a teacher encoder:
\begin{equation}
  \bar{f}_{t+1}^{\mathrm{tac}}
  =
  E_{\mathrm{tac}}^{\mathrm{EMA}}(I_{t+1}^{\mathrm{tac}}),
\end{equation}
where the teacher encoder $E_{\mathrm{tac}}^{\mathrm{EMA}}$ is updated as an exponential moving average (EMA) of $E_{\mathrm{tac}}$.
We use the InfoNCE contrastive objective~\cite{InfoNCE:Oord:arXiv2018}, which encourages the predicted feature to be similar to the corresponding target feature and dissimilar to other samples in the batch.
The tactile encoder is updated through this objective to capture task-specific temporal changes in contact states.
The mid-trained tactile encoder is then used to initialize the tactile encoder in Diffusion Policy for downstream imitation learning.


\begin{figure*}[tpb]
  \centering
  \setlength{\tabcolsep}{1pt}
  \begin{tabular}{ccccccc}
    \multicolumn{5}{c}{\footnotesize{\textbf{Real world}}} &
    \multicolumn{2}{c}{\footnotesize{\textbf{Simulation}}} \\[-1pt]

    \includegraphics[width=0.137\textwidth]{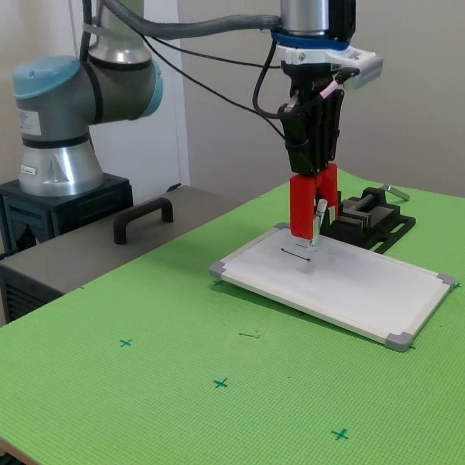} &
    \includegraphics[width=0.137\textwidth]{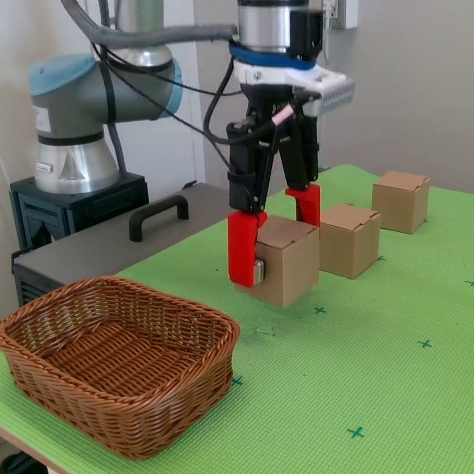} &
    \includegraphics[width=0.137\textwidth]{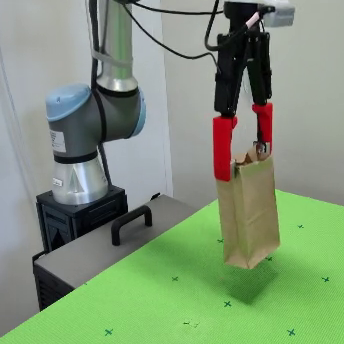} &
    \includegraphics[width=0.137\textwidth]{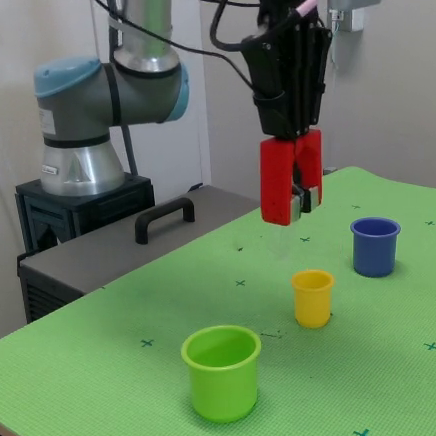} &
    \includegraphics[width=0.137\textwidth]{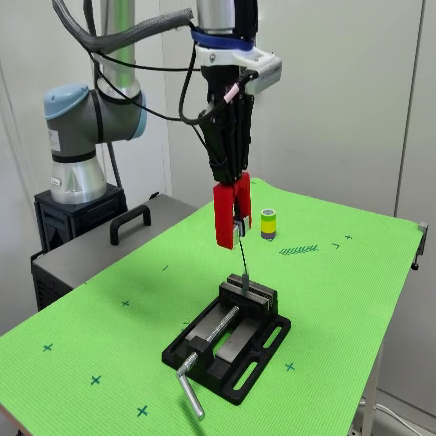} &
    \includegraphics[width=0.137\textwidth]{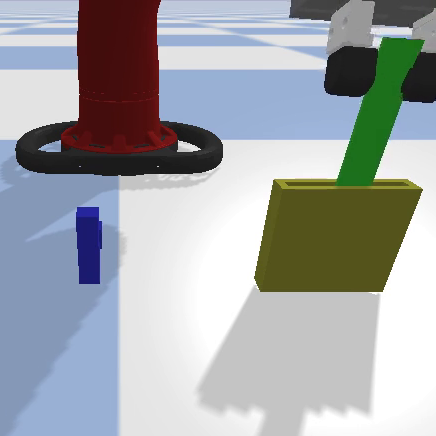} &
    \includegraphics[width=0.137\textwidth]{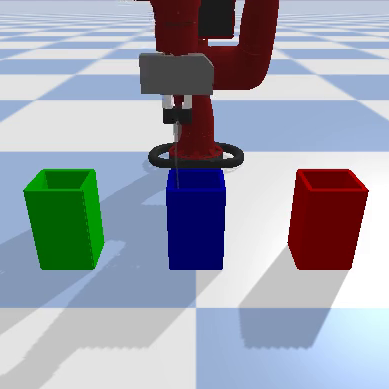} \\[-2pt]

    \includegraphics[width=0.137\textwidth]{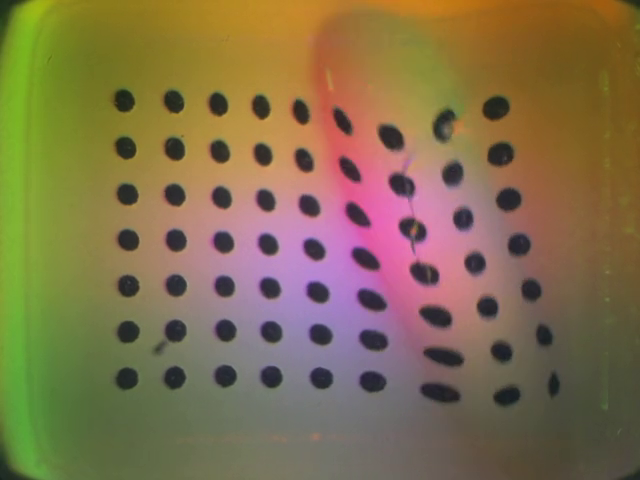} &
    \includegraphics[width=0.137\textwidth]{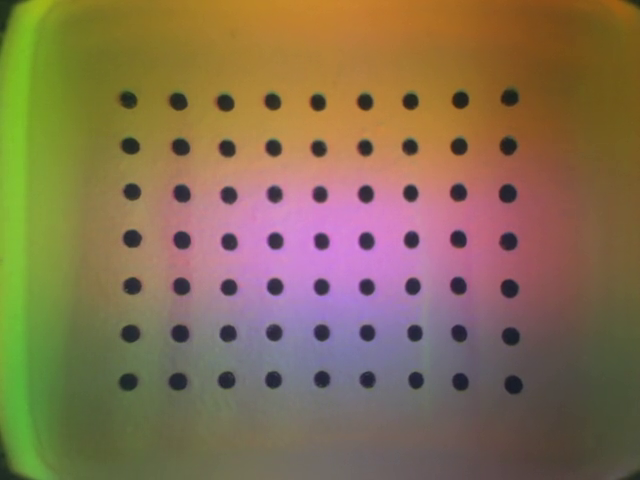} &
    \includegraphics[width=0.137\textwidth]{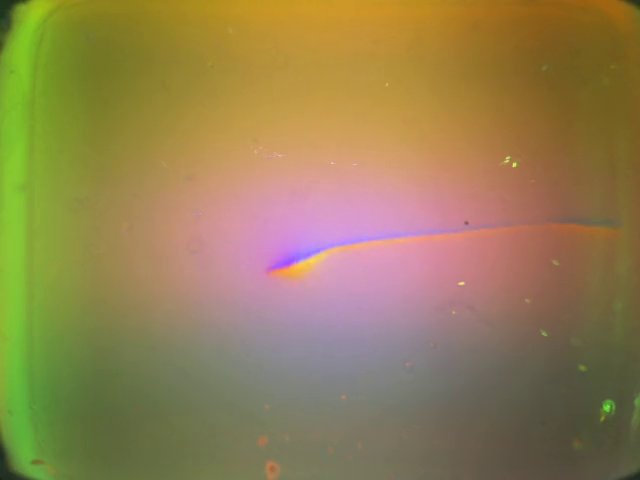} &
    \includegraphics[width=0.137\textwidth]{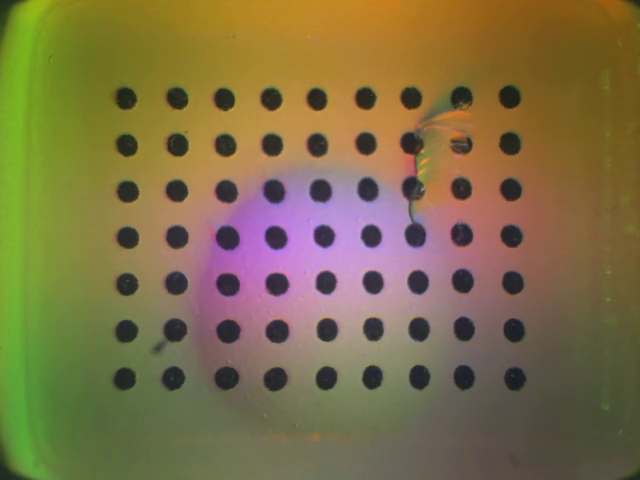} &
    \includegraphics[width=0.137\textwidth]{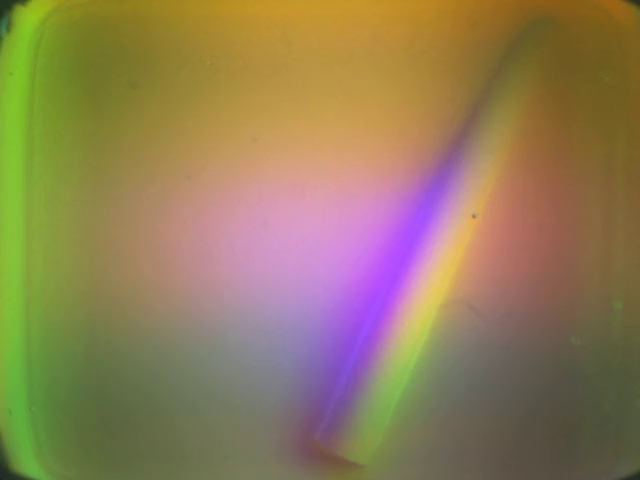} &
    \includegraphics[width=0.137\textwidth]{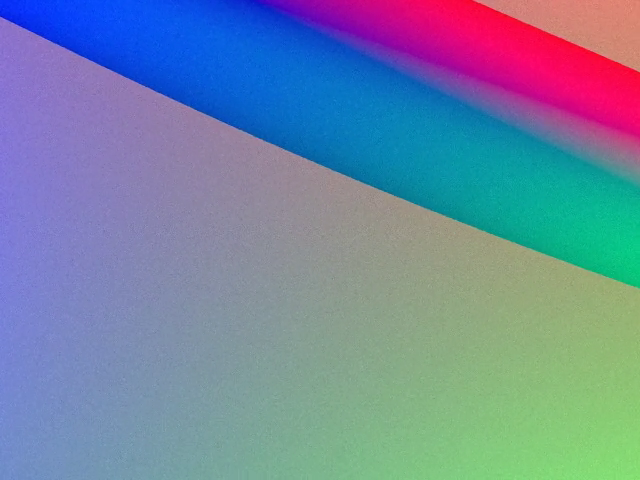} &
    \includegraphics[width=0.137\textwidth]{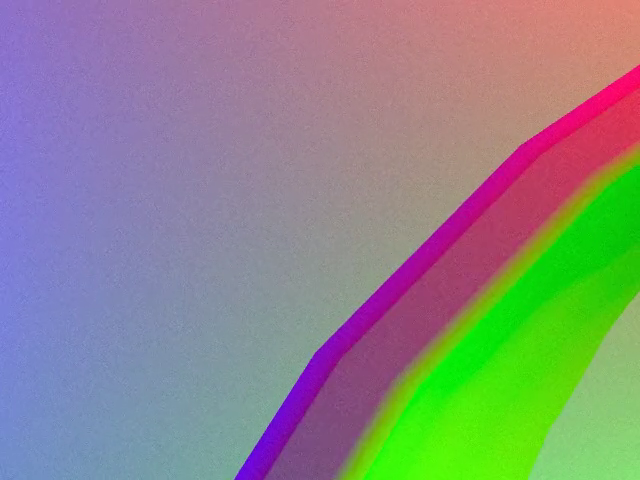} \\[-1pt]

    \scriptsize MarkerDrawing &
    \scriptsize HeavyBoxSorting &
    \scriptsize PaperBoxHandling &
    \scriptsize SpoonSorting &
    \scriptsize PenInsertion &
    \scriptsize PlateInsertion &
    \scriptsize SpoonSorting
  \end{tabular}
  \caption{
    Contact-rich manipulation tasks used in the experiments.
    \newline
    \footnotesize{
      The top and bottom rows show the front-view and tactile images, respectively.
      The five left columns correspond to real-world tasks, while the two right columns correspond to simulation tasks.
    }
  }
  \label{fig:tasks}
\end{figure*}

\section{Experiments} \label{sec:experiments}

\subsection{Experimental Setup}

We evaluate imitation learning with TFMs on contact-rich manipulation tasks in both real-world and simulation environments.


\subsubsection{Robot and Tactile Sensing Setup}

For the real-world experiments, we use a UR5e manipulator equipped with a Robotiq 2F-140 parallel gripper.
A GelSight Mini is mounted on the inner surface of each finger to capture tactile images, and a hand camera is mounted at the base of the gripper.
Both tactile and camera images are resized to $320\times240$ pixels before being provided to the policy.
The proprioceptive state consists of the joint angles of the 6-DoF arm and 1-DoF gripper, and the policy predicts the same joint angles as actions.

For the simulation experiments, we use TACTO~\cite{TACTO:Wang:RAL2022}, based on PyBullet, to simulate a Sawyer manipulator with a parallel gripper and optical tactile sensors.
TACTO provides tactile images in the same format as those obtained from the real-world GelSight sensors.
The other observations and actions follow the same setup as in the real-world experiments.

For both real-world and simulation experiments, we use RoboManipBaselines~\cite{RMB:Murooka:Access2026} for data collection, policy training, and deployment.



\subsubsection{Manipulation Tasks}

We consider several contact-rich manipulation tasks in which tactile information is important.

\textbf{MarkerDrawing:}
The robot grasps a marker and draws two parallel lines on a whiteboard.

\textbf{HeavyBoxSorting:}
The robot sequentially lifts three identical boxes, identifies the box containing an additional weight, and places the weighted box into a basket.

\textbf{PaperBoxHandling:}
The robot carefully grasps a deformable paper container without crushing it and moves it to a target location.

\textbf{SpoonSorting:}
The robot grasps a spoon and sorts it into one of two different cups depending on whether the handle or bowl side is grasped.

\textbf{PenInsertion:}
The robot grasps a ballpoint pen refill and inserts it into a pen barrel.

\textbf{PlateInsertion:}
The robot grasps a plate and inserts it into a narrow slit.

For the real-world experiments, we use subsets of these tasks depending on the evaluation, while the simulation experiments use PlateInsertion and SpoonSorting.
PenInsertion is evaluated only in the fine-tuning and mid-training experiments, not in the cross-TFM comparison.
The initial object poses are randomized for each rollout, and task success is determined manually.










\subsubsection{Demonstration Collection and Policy Training}

For each task, we collect 32 demonstrations through human teleoperation.
The operator commands relative end-effector motions using a 3D mouse, and joint-angle commands obtained through inverse kinematics are sent to the robot.
The same teleoperation procedure is used in both real-world and simulation environments, with observations and actions recorded every 100~ms.

All policies are trained for 480 epochs with a batch size of 32 and a learning rate of $1\times10^{-4}$.
We use an observation horizon of 2 and an action horizon of 8.
These training conditions are kept identical across all policies.
A separate policy is trained for each task, without multi-task learning.

Policy performance is evaluated by task success rate.
Our extensive real-world evaluation includes a total of 5,136 policy rollouts.

In the following experimental subsections, all policies are evaluated under identical conditions for each task.
However, some task objects differ across experiments.
Therefore, success rates should be compared within each subsection rather than across different subsections.




\begin{table*}[t]
  \centering
  \caption{Task success rates (\%) with different tactile representations in real-world and simulation experiments.}
  \label{tab:tfm}
  \begin{tabular}{l|ccccc|ccc}
    \toprule
     & \multicolumn{5}{c|}{Real world} & \multicolumn{3}{c}{Simulation} \\
    Policy
     & MarkerDrawing
     & HeavyBoxSorting
     & PaperBoxHandling
     & SpoonSorting
     & Avg.
     & PlateInsertion
     & SpoonSorting
     & Avg. \\
    \midrule
    DP-ResNetCamera
     & 8.3 & 47.2 & 47.2 & 55.6 & 39.6
     & 3.3 & 62.5 & 32.9 \\
    DP-ResNetTactile
     & 38.9 & 50.0 & 55.6 & 72.2 & 54.2
     & 66.7 & 87.5 & 77.1 \\
    \midrule
    DP-SparshMAE
     & 47.2 & 63.9 & 50.0 & 69.4 & 57.6
     & 51.7 & 95.8 & 73.8 \\
    DP-SparshDINO
     & 27.8 & 58.3 & 66.7 & 80.6 & 58.3
     & 41.7 & \textbf{100.0} & 70.8 \\
    DP-SparshDINOv2
     & \textbf{58.3} & \textbf{72.2} & 58.3 & 80.6 & \textbf{67.4}
     & 70.0 & \textbf{100.0} & 85.0 \\
    DP-SparshIJEPA
     & 33.3 & 38.9 & \textbf{72.2} & \textbf{83.3} & 56.9
     & 38.3 & 91.7 & 65.0 \\
    DP-AnyTouch
     & 41.7 & 50.0 & 47.2 & 52.8 & 47.9
     & 68.3 & 95.8 & 82.1 \\
    DP-T3
     & 47.2 & 36.1 & 44.4 & 66.7 & 48.6
     & \textbf{75.0} & \textbf{100.0} & \textbf{87.5} \\
    DP-UniT
     & 55.6 & 50.0 & 33.3 & 63.9 & 50.7
     & 71.7 & 91.7 & 81.7 \\
    \bottomrule
  \end{tabular}\\
  \vspace{2mm}
  \begin{minipage}{1.95\columnwidth}
    \footnotesize{
      Results are averaged over three training seeds.
      For each task, each training seed is evaluated with 12 and 20 rollouts in the real-world and simulation experiments, respectively.
    }
  \end{minipage}
\end{table*}

\subsection{Evaluation of Tactile Representations}

We evaluate the effect of pretrained tactile representations on imitation learning for contact-rich manipulation.
We compare \textbf{DP-ResNetCamera}, which encodes hand-camera images using ResNet-18, \textbf{DP-ResNetTactile}, which encodes tactile images using the same ResNet-18, and seven \textbf{DP-$\langle$\textit{TFM}$\rangle$} variants using the pretrained encoders described in Sec.~\ref{sec:method-tfm}.
All pretrained tactile encoders are fine-tuned during downstream imitation learning.


\subsubsection{Real-World Experiments}

The real-world results are shown in \tabref{tab:tfm}.
The average success rate across four tasks increases from 39.6\% with DP-ResNetCamera to 54.2\% with DP-ResNetTactile, demonstrating the benefit of tactile observations for contact-rich manipulation.
Policies using pretrained tactile representations achieve average success rates ranging from 47.9\% to 67.4\%, showing substantial performance differences across representations.
Among them, DP-SparshDINOv2 achieves the highest average success rate, outperforming DP-ResNetTactile by 13.2 percentage points and achieving higher success rates on all four tasks.
These results show that an appropriate pretrained tactile representation can outperform a generic visual encoder designed for RGB images when applied to tactile images.
However, not all pretrained tactile representations outperform DP-ResNetTactile, indicating that pretraining alone does not guarantee consistent performance improvements.


\subsubsection{Simulation Experiments}

The simulation results are also shown in \tabref{tab:tfm}.
The average success rate across two tasks increases from 32.9\% with DP-ResNetCamera to 77.1\% with DP-ResNetTactile, again showing a substantial benefit from tactile observations.
Most policies using pretrained tactile representations perform comparably to or better than DP-ResNetTactile.
DP-T3 and DP-SparshDINOv2 achieve particularly high average success rates of 87.5\% and 85.0\%, respectively.
However, the performance trends across pretrained tactile representations are not necessarily consistent between simulation and the real world.
This discrepancy may be attributed to the domain gap between simulated and real-world tactile images.
The robustness of pretrained tactile representations to such domain gaps remains an important direction for future investigation.


\begin{figure}[tpb]
  \centering
  \includegraphics[width=1.0\columnwidth]{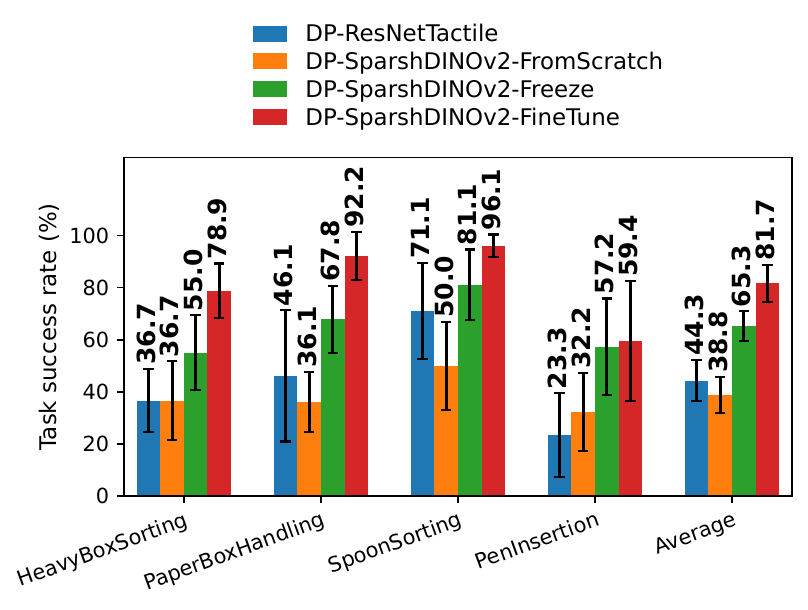}
  \caption{
    Effect of fine-tuning.
    \newline
    \footnotesize{
      Each bar shows the mean task success rate over 15 training seeds, with 12 evaluation rollouts per seed.
      Error bars indicate the standard deviation across training seeds.
      Average denotes the mean success rate across the four real-world tasks.
    }
  }
  \label{fig:adaptation}
\end{figure}

\subsection{Effect of Fine-Tuning}

As described in Sec.~\ref{sec:method-adaptation}, we investigate how pretrained tactile representations should be adapted to downstream imitation learning.
For this evaluation, we use Sparsh DINOv2, which achieved the highest average success rate in the previous subsection.
We compare \textbf{DP-SparshDINOv2-FromScratch}, which trains the encoder without pretrained weights, \textbf{DP-SparshDINOv2-Freeze}, which keeps the pretrained encoder fixed, and \textbf{DP-SparshDINOv2-FineTune}, which updates the pretrained encoder together with the policy.
We also include \textbf{DP-ResNetTactile} as a baseline without a TFM.

The results are shown in \figref{fig:adaptation}.
The average success rates across four real-world tasks are 44.3\% for DP-ResNetTactile and 38.8\% for DP-SparshDINOv2-FromScratch, compared with 65.3\% for DP-SparshDINOv2-Freeze and 81.7\% for DP-SparshDINOv2-FineTune.
While DP-SparshDINOv2-FromScratch does not outperform DP-ResNetTactile, both DP-SparshDINOv2-Freeze and DP-SparshDINOv2-FineTune achieve substantially higher performance, demonstrating the benefit of Sparsh DINOv2 pretraining.
Furthermore, DP-SparshDINOv2-FineTune outperforms DP-SparshDINOv2-Freeze by 16.4 percentage points on average and achieves the highest success rate on all four tasks.
For the ViT-based Sparsh DINOv2 encoder, these results demonstrate the benefits of tactile-image pretraining over random initialization and of fine-tuning over freezing.
Accordingly, for the tasks studied here, we recommend selecting a high-performing pretrained TFM (Sparsh DINOv2 in our comparison) and fine-tuning it for the target task.
Whether the same fine-tuning benefit holds across other TFMs and architectures remains to be investigated.



\begin{figure}[tpb]
  \centering
  \includegraphics[width=0.92\columnwidth]{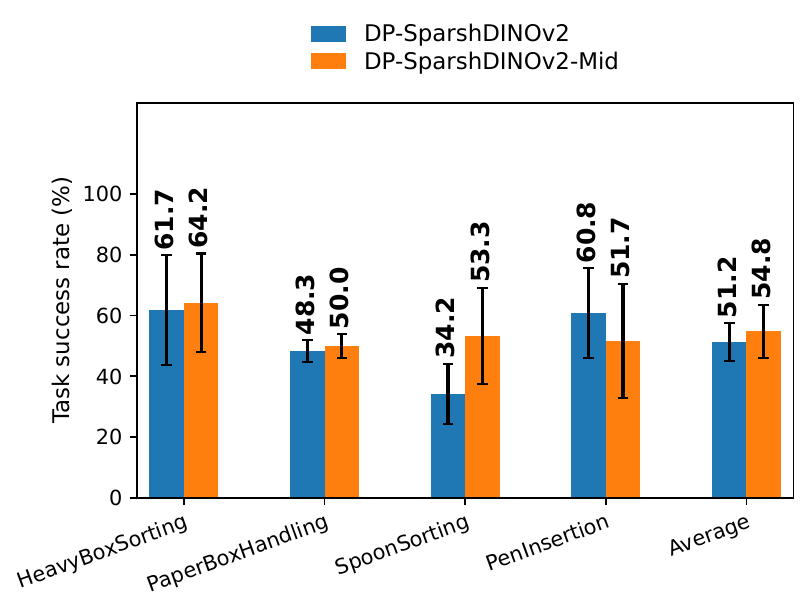}
  \caption{
    Effect of task-specific mid-training.
    \newline
    \footnotesize{
      Each bar shows the mean task success rate over 10 training seeds, with 12 evaluation rollouts per seed.
      Error bars indicate the standard deviation across training seeds.
      Average denotes the mean success rate across the four real-world tasks.
    }
  }  \label{fig:midtraining}
\end{figure}

\subsection{Effect of Task-Specific Mid-Training}

As described in Sec.~\ref{sec:method-mid-training}, we evaluate the effect of task-specific mid-training on downstream imitation learning.
We compare \textbf{DP-SparshDINOv2}, which directly fine-tunes the pretrained encoder, with \textbf{DP-SparshDINOv2-Mid}, which first performs task-specific mid-training and then fine-tunes the resulting encoder.

The results are shown in \figref{fig:midtraining}.
The average success rate across four real-world tasks slightly increases from 51.2\% without mid-training to 54.8\% with mid-training.
At the task level, mid-training improves performance on HeavyBoxSorting, SpoonSorting, and PaperBoxHandling, while reducing performance on PenInsertion.
These results indicate that task-specific mid-training can improve downstream imitation learning performance, but its effectiveness depends on the task.



\section{Conclusion}

In this work, we systematically evaluated multiple TFMs integrated into Diffusion Policy for end-to-end imitation learning of contact-rich manipulation in both real-world and simulation environments.
Our experiments showed that appropriate pretrained tactile representations can outperform a generic visual encoder applied to tactile images, demonstrating the effectiveness of TFMs for imitation learning.
At the same time, the performance varied substantially across pretrained tactile representations.
For Sparsh DINOv2, fine-tuning outperformed freezing the pretrained encoder, whereas the benefit of mid-training was task-dependent.
These findings support selecting a promising pretrained TFM and fine-tuning it for the target task, although the effectiveness of this procedure for other TFMs remains to be established.

Enabling robots to acquire general-purpose prior knowledge of tactile interactions before learning downstream manipulation tasks is an important direction for tactile robot learning.
Building on our findings, future work will explore methods for selecting suitable pretrained tactile representations from task-specific data and more effective mid-training strategies for adapting these representations to downstream imitation learning.



\addtolength{\textheight}{-10.0cm}  

\bibliographystyle{IEEEtran}
\bibliography{main.bib}

\end{document}